\documentclass[runningheads]{llncs}
\usepackage[T1]{fontenc}
\usepackage{amsmath}
\usepackage{amssymb}
\usepackage{xcolor}
\usepackage{graphicx}
\usepackage{enumitem}

\newcommand{\mammolife}{Mammo-LIFE}
\begin{document}

\title{\mammolife: Longitudinal Mammographic Imaging and Clinical Feature Enrichment for Post-Radiotherapy Outcome Prediction}
%
%
\titlerunning{Mammo-LIFE for Radiotherapy Outcome Prediction}

%
%
%
%
\author{Farnoush Bayatmakou\inst{1} \and
Maryam Hosseini\inst{1} \and
Reza Taleei\inst{2} \and
Arash~Mohammadi\inst{1}}

\authorrunning{F. Bayatmakou et al.}
%
\institute{Department of Cybersecurity and Intelligent Systems Engineering (CISE),
Concordia University, Montreal, QC, Canada\\
\email{farnoush.bayatmakou@mail.concordia.ca,
arash.mohammadi@concordia.ca}
\and
Department of Radiation Oncology, Atrium Health Levine Cancer Institute,
Charlotte, NC, USA}
\maketitle
%
\begin{abstract}
Recent advances in Artificial Intelligence (AI)-powered Computer-Aided Diagnosis (CAD) systems have substantially improved breast cancer screening, diagnosis, and prognosis. Comparatively, post-radiotherapy outcome prediction using paired longitudinal mammograms has received considerably less attention. This is largely due to the limited availability of well-annotated longitudinal datasets. Longitudinal mammograms, coupled with paired pre- and post-treatment information, provide a unique opportunity to characterize treatment-induced breast tissue changes following radiotherapy. The resulting learned representations can serve as a valuable asset for advancing personalized radiotherapy planning and post-treatment management. In this context, we propose \mammolife, a patient-level multimodal framework for post-radiotherapy outcome prediction that combines longitudinal mammographic features with patient-level clinical variables. The imaging branch processes paired pre- and post-treatment mammograms acquired from the four standard views using a mammography-specific encoder adapted via Low-Rank Adaptation (LoRA). Within each view, pre- and post-treatment representations are explicitly compared through a longitudinal comparison module to capture treatment-related changes. The resulting view-level embeddings are then aggregated using learned view-attention pooling to form a unified patient-level mammographic representation. Selected clinical variables are subsequently combined with the image-derived prediction probability through a late-fusion strategy. To evaluate the effectiveness of combining paired longitudinal mammograms with clinical information, experiments were conducted on an in-house clinical cohort using patient-level stratified five-fold cross-validation. The proposed \mammolife{} demonstrated strong cross-validated performance across the evaluated configurations. The best setting achieved an AUC of $0.86 \pm 0.14$, accuracy of $0.79 \pm 0.15$, and F1-score of $0.83 \pm 0.11$. 
\end{abstract}
%
\section{Introduction}
Mammography plays a central role in breast cancer screening, diagnosis, and post-treatment follow-up~\cite{niell2024acr}. Standard mammographic examinations typically include bilateral craniocaudal (CC) and mediolateral oblique (MLO) views, providing complementary projections for clinical interpretation~\cite{monticciolo2023breast}. Beyond screening and diagnosis, mammography supports post-treatment follow-up by providing longitudinal imaging records of breast tissue changes over time~\cite{swinnen2018breast}. During post-treatment follow-up, mammographic appearances are influenced by the type of therapy received. Accordingly, follow-up mammograms after radiotherapy may exhibit expected benign findings such as edema-related thickening, postoperative scarring, architectural distortion, and other treatment-related changes~\cite{chansakul2012postconservation}. These findings evolve over time and vary across patients, making paired pre- and post-treatment mammograms a natural basis for longitudinal analysis.

Modeling longitudinal mammographic information requires robust representation learning across views and time points. Deep learning approaches have been increasingly explored for mammography-based screening and cancer detection~\cite{wu2019deep,mckinney2020international}, including multi-view convolutional, graph-, and transformer-based architectures that jointly exploit the standard mammographic views~\cite{geras2017high,manigrasso2025mammography,sun2022transformer}. More recently, mammography-specific foundation and vision-language models, such as Mammo-CLIP~\cite{ghosh2024mammoclip} and VersaMammo~\cite{huang2025versamammo}, have demonstrated transferable representations for downstream breast imaging tasks. Longitudinal mammography has also been investigated for breast cancer risk prediction~\cite{dadsetan2022deep} and diagnosis using Mamba-based state-space modeling~\cite{zhou2025longitudinalmamba}. Beyond imaging, routinely collected clinical variables, including tumor size, histologic grade, hormone receptor status, and body mass index, provide complementary patient-level prognostic information~\cite{loibl2024early,chan2014body}, motivating multimodal prediction frameworks. Despite recent advances in foundation model adaptation for mammographic image analysis~\cite{bayatmakou2026fewshot,huang2025versamammo,ghosh2024mammoclip,ghosh2025mammo,zhou2026mammodino}, relatively few studies have investigated longitudinal mammographic modeling. Among those using paired pre/post-treatment mammograms, only one has explored multi-view post-radiotherapy analysis, and it did not incorporate patient-level clinical variables. Consequently, post-radiotherapy outcome prediction using paired longitudinal mammograms remains relatively underexplored.
\begin{figure}[t]
    \centering
\includegraphics[width=.95\linewidth,
keepaspectratio]{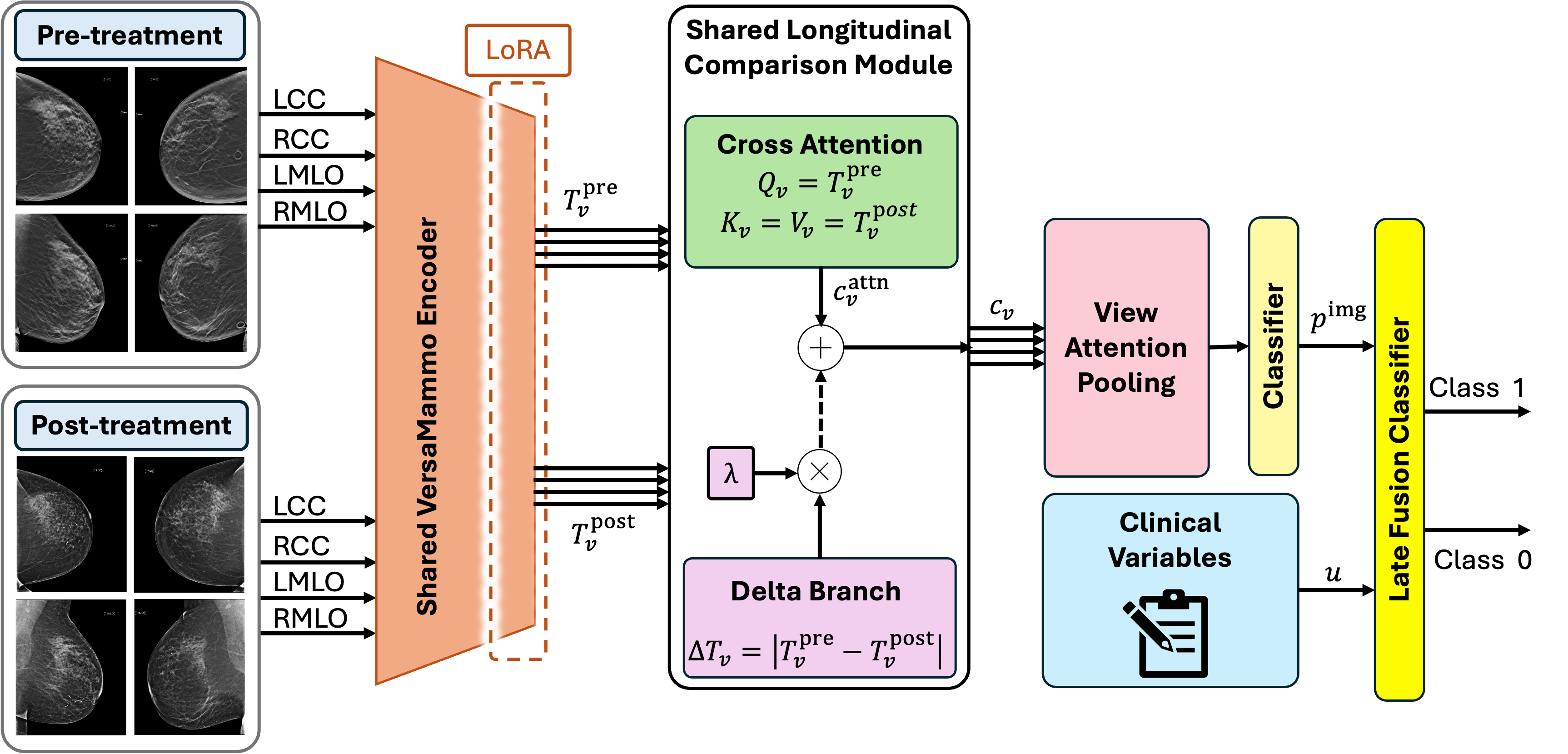}
    \caption{Overview of \mammolife{}. Paired pre/post mammograms are encoded using a LoRA-adapted VersaMammo backbone and compared using either a cross-attention-only or a delta-enhanced longitudinal design. View-level embeddings are aggregated through view-attention pooling, and the image-derived prediction probability is combined with selected clinical variables through late fusion for final outcome prediction.}
    \label{fig:framework}
\end{figure}
Motivated by the above research gap, we propose \mammolife, a patient-level multimodal framework that combines paired pre- and post-treatment mammograms with selected clinical variables for post-radiotherapy outcome prediction. \mammolife{} builds upon a Low-Rank Adaptation (LoRA)-adapted EfficientNet-B5 variant of VersaMammo to learn longitudinal mammographic representations from paired pre- and post-treatment images. The imaging branch compares pre- and post-treatment representations across the four standard mammographic views and aggregates the resulting view-level embeddings using learned view-attention pooling. Clinical variables are incorporated via late fusion by combining the image-derived prediction probability with selected patient-level clinical features. Unlike the prior multi-view longitudinal pre/post mammogram analysis approach, which used a general vision foundation model in a few-shot evaluation setting, \mammolife{} employs a LoRA-adapted mammography-specific VersaMammo backbone, incorporates selected clinical variables through late fusion, and is evaluated under patient-level stratified 5-fold cross-validation. Its longitudinal pre/post comparison stage considers both a cross-attention-only design and a delta-enhanced cross-attention variant, where explicit pre/post temporal-difference features augment the cross-attention representation. In brief, the main contributions of this work are summarized as follows:
\begin{itemize}[label=$\bullet$]
\item We present \mammolife, a patient-level multimodal framework for post-radiotherapy outcome prediction that jointly exploits paired pre- and post-treatment mammograms together with selected clinical variables, addressing a relatively underexplored problem in breast imaging.
\item We formulate post-radiotherapy outcome prediction as a patient-level longitudinal representation learning problem by jointly modeling paired pre- and post-treatment mammograms acquired from the four standard mammographic views. To this end, we develop a parameter-efficient adaptation framework based on a LoRA-adapted mammography-specific VersaMammo.
\item We introduce a longitudinal feature comparison module that investigates both cross-attention and delta-enhanced cross-attention to model treatment-induced changes between paired pre- and post-treatment mammograms, followed by learned view-attention pooling for patient-level representation learning.
\item We demonstrate that integrating selected clinical variables with longitudinal mammographic representations through late multimodal fusion improves post-radiotherapy outcome prediction over image-only modeling, highlighting the complementary value of imaging and clinical information.
\end{itemize}

\section{Methodology} \label{sec:proposed_framework}
This section presents \mammolife, illustrated in Figure~\ref{fig:framework}, which consists of four main components: LoRA-adapted VersaMammo feature extraction, longitudinal pre/post comparison, view-attention pooling, and late clinical fusion.

\textbf{\textit{VersaMammo Backbone and LoRA Adaptation:}}
The imaging encoder is initialized with the EfficientNet-B5 variant of VersaMammo~\cite{huang2025versamammo}, a mammography-specific foundation model developed for AI-enabled mammogram interpretation. VersaMammo follows a two-stage pretraining strategy, in which a teacher model is first trained using self-supervised learning on unlabeled mammograms, and supervised learning with knowledge distillation is then used to transfer learned representations and clinical knowledge to the student model, thereby enabling domain-specific mammographic feature extraction. To adapt the pretrained encoder to post-radiotherapy outcome prediction, we apply LoRA~\cite{hu2022lora} to selected later-stage convolutional layers of the EfficientNet-B5 backbone. Let $E_{\theta}$ denote the encoder with frozen pretrained parameters $\theta$. During training, only the LoRA parameters and task-specific layers, including the feature-to-token projection, longitudinal comparison, view-attention pooling, and classification modules, are optimized for parameter-efficient adaptation.

\textbf{\textit{Patient-Level Longitudinal Input Representation:}}
The imaging input consists of paired pre- and post-treatment mammograms. At each time point, a set of standard images corresponding to left and right sides $\{L, R\}$ and projections $\{CC, MLO\}$ is considered. Let $x^{\mathrm{pre}}_{i,v}$ and $x^{\mathrm{post}}_{i,v}$, respectively, denote the pre/post-treatment images of patient $i$ from view \[v\in V:=\{\mathrm{LCC}, \mathrm{LMLO}, \mathrm{RCC}, \mathrm{RMLO}\}.\] Subsequently, the complete patient-level imaging input is defined as
\[
X_i =
\left\{
\left(x^{\mathrm{pre}}_{i,v}, x^{\mathrm{post}}_{i,v}\right)
\mid
v\in V
\right\}.
\]
Each patient is associated with a binary outcome label $y_i \in \{0,1\}$ derived from an ordinal radiologist-assigned assessment of density- and fibrosis-related changes between the pre- and post-treatment mammograms. The model is trained at the patient level, using all four pre/post view pairs to predict $y_i$.

\textbf{\textit{Adapted Spatial Feature Extraction:}}
Given the patient-level imaging input $X_i$, each mammogram is processed using a shared LoRA-adapted VersaMammo encoder. The same encoder is applied across all views and time points to obtain feature maps in a shared feature space. For patient $i$ and view $v$, the paired feature maps are computed as
\[
F^{\mathrm{pre}}_{i,v} =
E_{\theta,\phi}\left(x^{\mathrm{pre}}_{i,v}\right),
\qquad
F^{\mathrm{post}}_{i,v} =
E_{\theta,\phi}\left(x^{\mathrm{post}}_{i,v}\right),
\]
where $E_{\theta,\phi}$ denotes the encoder with frozen pretrained parameters $\theta$ and trainable LoRA parameters $\phi$. Each extracted feature map $F \in \mathbb{R}^{D \times H' \times W'}$ is reshaped into $N=H'W'$ spatial tokens and projected to a common embedding dimension $d$ for longitudinal comparison. The resulting token sequences,
$T^{\mathrm{pre}}_{i,v}, T^{\mathrm{post}}_{i,v} \in \mathbb{R}^{N \times d}$, are used to model view-specific changes over time.

\textbf{\textit{Longitudinal Pre/Post Feature Comparison:}}
To model treatment-related longitudinal changes, the pre/post-treatment token sequences are compared within each standard view. We evaluate two designs for the longitudinal comparison module before view-attention aggregation. The baseline design uses cross-attention, where pre-treatment tokens serve as queries and post-treatment tokens serve as keys and values. The attention output is added to the pre-treatment tokens and refined using a residual feed-forward block. The attention-based view-level embedding is obtained by mean pooling as
\begin{eqnarray}
H_{i,v}
&=&
T^{\mathrm{pre}}_{i,v}
+
\mathrm{CrossAttn}
\left(
T^{\mathrm{pre}}_{i,v},
T^{\mathrm{post}}_{i,v},
T^{\mathrm{post}}_{i,v}
\right),\nonumber\\
c^{\mathrm{attn}}_{i,v}
&=&
\mathrm{MeanPool}
\big(
\mathrm{FFN}_{\mathrm{res}}
(
H_{i,v}
)
\big),\nonumber
\end{eqnarray}
where $\mathrm{FFN}_{\mathrm{res}}$ denotes the residual feed-forward refinement block. To assess whether explicit temporal-difference cues provide complementary information, we evaluate a delta-enhanced design using the token-wise absolute difference as
\[
\Delta T_{i,v}
=
\left|
T^{\mathrm{post}}_{i,v}
-
T^{\mathrm{pre}}_{i,v}
\right|.
\]
In this variant, the attention-refined tokens and delta tokens are first mean-pooled, and the resulting view-level representations are combined as
\[
c_{i,v}
=
c^{\mathrm{attn}}_{i,v}
+
\lambda \,
\mathrm{MeanPool}
\left(
\Delta T_{i,v}
\right),
\]
where $\lambda$ is a learnable scalar parameter.
For the cross-attention-only design, the final view-level embedding is $c_{i,v}=c^{\mathrm{attn}}_{i,v}$. This produces one view-specific longitudinal embedding for each standard view.

\textbf{\textit{View-Attention Pooling for Patient-Level Aggregation:}}
After view-level pre/post comparison, each patient is represented by four view-specific change embeddings. These embeddings are stacked as
\begin{equation}
    C_i =
\left[
c_{i,\mathrm{LCC}}\quad
c_{i,\mathrm{LMLO}}\quad
c_{i,\mathrm{RCC}}\quad
c_{i,\mathrm{RMLO}}
\right]^{\top}
\in \mathbb{R}^{4 \times d}. \notag
\end{equation}
To aggregate the view-specific embeddings into a unified patient-level mammographic representation, we use learned view-attention pooling. For each view-level embedding, an attention score is computed as
\[
a_{i,v}
=
w^\top c_{i,v} + b,
\qquad
\forall v \in V,
\]
where $w$ and $b$ are learnable parameters. The patient-level longitudinal mammographic representation is then obtained as
\[
z^{\mathrm{img}}_i
=
\sum_{v \in V}
\alpha_{i,v} c_{i,v},
\]
where $\alpha_{i,v}
=
\frac{
\exp(a_{i,v})
}{
\sum_{v' \in V}
\exp(a_{i,v'})
}$. The representation $z^{\mathrm{img}}_i$ is passed to an image classification head followed by a sigmoid activation to obtain the image-derived probability $p^{\mathrm{img}}_i$. Under patient-level stratified cross-validation, each patient is assigned an out-of-fold image-derived probability based on the corresponding held-out fold. This probability summarizes the longitudinal imaging evidence from the paired mammograms.

\textbf{\textit{Clinical Feature Processing and Late Multimodal Fusion:}}
In addition to the longitudinal mammographic representation, we incorporate selected patient-level clinical variables as complementary non-imaging information. Clinical variables are integrated through prediction-level late fusion. Specifically, the image-derived prediction probability $p^{\mathrm{img}}_i$ is concatenated with the selected clinical feature vector $u_i$ to form the late-fusion representation 
$[
p^{\mathrm{img}}_i ;
u_i
]$. The combined representation is used to train the final prediction model using only the training patients within each fold, and performance is evaluated on the corresponding held-out patients. The same patient-level folds are used for image-only and image-plus-clinical evaluation, enabling patient-matched comparison while avoiding patient-level leakage.
\section{Experiments and Results}
\textbf{\textit{Dataset and Outcome Definition:}}
We evaluate \mammolife{} on an in-house clinical cohort consisting of $47$ patients with paired pre- and post-radiotherapy multi-view mammograms from the four standard views, along with associated patient-level clinical variables. Each patient is assigned an ordinal radiologist score indicating the perceived direction of density- and fibrosis-related change between the paired examinations. For binary prediction, scores indicating greater density/fibrosis on the pre-treatment mammographic examination are assigned to Class 0, corresponding to a perceived decrease after treatment. Scores indicating no perceived difference or greater density/fibrosis on the post-treatment mammographic examination are assigned to Class 1, corresponding to no appreciable change or a perceived increase. The ordinal score is used only to derive the binary outcome label and is not included as an input feature. All experiments are conducted using patient-level cross-validation, ensuring that all images and clinical variables associated with the same patient remain within the same fold, thereby preventing data leakage during evaluation.

\textbf{\textit{Implementation Details:}}
We evaluate two patient-level prediction settings: image-only and image-plus-clinical late fusion. All mammograms are resized to $224 \times 224$ pixels. The imaging backbone is initialized with the EfficientNet-B5 variant of VersaMammo, while pretrained backbone weights remain frozen. LoRA adapters are applied to selected later-stage $1\times1$ convolutional layers of the backbone with rank $r=4$, scaling factor $\alpha=4$, and LoRA dropout $0.10$.
Feature maps are reshaped into token sequences and projected to an embedding dimension of $d=512$. The longitudinal comparison module uses 8-head cross-attention with feed-forward refinement and mean pooling. The image model is trained for up to $30$ epochs using AdamW with learning rate $2\times10^{-5}$ and weight decay $3\times10^{-3}$.
In the image-plus-clinical setting, clinical variables are processed at the patient level, and the out-of-fold image-derived probability is concatenated with the selected clinical feature vector to form the late-fusion representation. Late fusion is implemented using logistic regression and random forest (RF) classifiers.

\textbf{\textit{Evaluation Protocol:}}
All experiments use patient-level stratified 5-fold cross-validation. Within each outer fold, the training patients are further split into training and validation subsets for checkpoint selection and early stopping. Early stopping is performed with a patience of $10$ epochs, and checkpoints are selected primarily by validation AUC, with validation loss used as a secondary criterion. The same outer folds are used across all evaluations.

\textbf{\textit{Results:}}
We first assess the effect of parameter-efficient adaptation within the imaging branch of \mammolife{}. This ablation is performed using the cross-attention longitudinal design. Table~\ref{tab:lora_ablation} compares frozen and LoRA-adapted VersaMammo backbones under this setting. The LoRA-adapted backbone improved image-only cross-attention performance relative to the frozen backbone, yielding higher AUC, accuracy, sensitivity, and F1-score. Specificity was slightly higher for the frozen backbone, but showed greater variability across folds. We then compare two longitudinal modeling designs within the \mammolife{} framework using the LoRA-adapted VersaMammo backbone. Table~\ref{tab:image_only_variants} compares the cross-attention design with the delta-enhanced design using mammographic inputs only. The delta-enhanced design slightly improved mean AUC, while threshold-based metrics remained similar.

\begin{table}[!t]
\caption{Effect of LoRA adaptation on the image-only cross-attention variant. Values are mean $\pm$ standard deviation across five folds.}
\label{tab:lora_ablation}
\centering
\footnotesize
\setlength{\tabcolsep}{2pt}
\begin{tabular}{|l|c|c|c|c|c|}
\hline
\textbf{Adaptation} & \textbf{AUC} & \textbf{Acc.} & \textbf{Sens.} & \textbf{Spec.} & \textbf{F1} \\
\hline
Frozen
& $0.64 \pm 0.13$
& $0.57 \pm 0.18$
& $0.57 \pm 0.44$
& $\mathbf{0.65 \pm 0.34}$
& $0.53 \pm 0.33$ \\
Adapted
& $\mathbf{0.72 \pm 0.13}$
& $\mathbf{0.70 \pm 0.18}$
& $\mathbf{0.76 \pm 0.25}$
& $0.62{\pm}0.21$
& $\mathbf{0.74 \pm 0.18}$ \\
\hline
\end{tabular}
\end{table}
\begin{table}[t]
\caption{Image-only performance across longitudinal designs using the LoRA-adapted backbone. Values are mean $\pm$ standard deviation across five folds.}
\label{tab:image_only_variants}
\centering
\footnotesize
\setlength{\tabcolsep}{5pt}
\begin{tabular}{|l|c|c|c|c|c|}
\hline
\textbf{Design} & \textbf{AUC} & \textbf{Acc.} & \textbf{Sens.} & \textbf{Spec.} & \textbf{F1} \\
\hline
Cross-Attn.
& $0.72\pm0.13$
& $0.70\pm0.18$
& $0.76\pm0.25$
& $0.62\pm0.21$
& $0.74\pm0.18$ \\
Delta-Enh.
& $\mathbf{0.73\pm0.11}$
& $0.70\pm0.18$
& $0.76\pm0.25$
& $0.62\pm0.21$
& $0.74\pm0.10$ \\
\hline
\end{tabular}
\end{table}
%
\begin{table}[t]
\caption{Image-plus-clinical late-fusion performance with the selected Tumor-Metabolic subset. Values are mean $\pm$ standard deviation across five folds.}
\label{tab:fusion_results}
\centering
\footnotesize
\setlength{\tabcolsep}{2.5pt}
\begin{tabular}{|l|l|c|c|c|c|c|}
\hline
\textbf{Design} & \textbf{Clf.} & \textbf{AUC} & \textbf{Acc.} & \textbf{Sens.} & \textbf{Spec.} & \textbf{F1} \\
\hline
Cross-Attn. & RF
& $\mathbf{0.86{\pm}0.14}$
& $0.77{{\pm}} 0.11$
& $\mathbf{0.83{\pm}0.12}$
& $0.67{\pm} 0.20$
& $0.82{\pm} 0.09$ \\
Delta-Enh. & RF
& $\mathbf{0.86{\pm}0.14}$
& $\mathbf{0.79{\pm}0.15}$
& $\mathbf{0.83{\pm}0.12}$
& $\mathbf{0.73{\pm}0.25}$
& $\mathbf{0.83{\pm}0.11}$ \\
Cross-Attn. & Log. Reg.
& $0.81{\pm}0.16$
& $0.75{\pm}0.14$
& $0.77{\pm}0.15$
& $0.72{\pm}0.18$
& $0.79{\pm}0.12$ \\
Delta-Enh. & Log. Reg.
& $0.81{\pm}0.16$
& $0.75{\pm}0.14$
& $0.77{\pm}0.15$
& $0.72{\pm}0.18$
& $0.79{\pm}0.12$ \\
\hline
\end{tabular}
\end{table}

We next evaluate image-plus-clinical late fusion by combining the image-derived probability with selected patient-level clinical variables. We considered clinically motivated feature subsets reflecting tumor characteristics, receptor status, metabolic factors, and age-related information, while preserving the same patient-level folds across all configurations. The Tumor-Metabolic subset showed the highest cross-validated performance among the tested subsets and was used for the main fusion comparison. Table~\ref{tab:fusion_results} reports performance for both longitudinal designs using this subset with logistic regression and random forest classifiers. Random forest achieved higher mean accuracy and F1-score than logistic regression. The delta-enhanced design with random forest achieved the highest accuracy, specificity, and F1-score, while tying with cross-attention using random forest for the highest AUC. Specifically, it achieved an AUC of $0.86 \pm 0.14$, accuracy of $0.79 \pm 0.15$, sensitivity of $0.83 \pm 0.12$, specificity of $0.73 \pm 0.25$, and F1-score of $0.83 \pm 0.11$. To assess the predictive value of clinical variables alone, we evaluated a clinical-only baseline using a random forest classifier with the same Tumor-Metabolic subset. This baseline achieved an AUC of $0.73 \pm 0.09$, accuracy of $0.62 \pm 0.09$, sensitivity of $0.69 \pm 0.30$, specificity of $0.50 \pm 0.40$, and F1-score of $0.66 \pm 0.14$. Compared with both image-only and clinical-only prediction, image-plus-clinical late fusion achieved stronger overall performance, suggesting complementary value from the two modalities in this cohort.
 
To further examine the contribution of different clinical variable groups, Table~\ref{tab:clinical_subset_analysis} compares four feature subsets within the delta-enhanced image-plus-clinical model with a random forest classifier: Tumor-Metabolic (Size, Grade, body mass index [BMI], Glucose, Diabetes), Tumor-Receptor (Size, Grade, estrogen receptor [ER], progesterone receptor [PR]), Receptor-Metabolic (ER, PR, BMI, Glucose, Diabetes, hypertension [HTN]), and Tumor-Receptor-Age (Age at diagnosis, Size, Grade, ER, PR). Tumor-Metabolic showed the highest overall performance.
\begin{table}[t]
\caption{Clinical subset analysis for the delta-enhanced image-plus-clinical model with a random forest classifier. Values are mean $\pm$ standard deviation across five folds.}
\label{tab:clinical_subset_analysis}
\centering
\footnotesize
\setlength{\tabcolsep}{2pt}
\begin{tabular}{|l|c|c|c|c|c|}
\hline
\textbf{Clinical subset}
& \textbf{AUC}
& \textbf{Acc.}
& \textbf{Sens.}
& \textbf{Spec.}
& \textbf{F1} \\
\hline
Tumor-Metabolic
& \textbf{0.86$\pm$0.14}
& \textbf{0.79$\pm$0.15}
& \textbf{0.83$\pm$0.12}
& \textbf{0.73$\pm$0.25}
& \textbf{0.83$\pm$0.11} \\
\hline
Tumor-Receptor
& 0.83$\pm$0.18
& 0.64$\pm$0.16
& 0.66$\pm$0.28
& 0.63$\pm$0.38
& 0.67$\pm$0.23 \\
\hline
Receptor-Metabolic
& 0.78$\pm$0.11
& 0.66$\pm$0.14
& 0.69$\pm$0.32
& 0.63$\pm$0.28
& 0.68$\pm$0.23 \\
\hline
Tumor-Receptor-Age
& 0.78$\pm$0.18
& 0.69$\pm$0.22
& 0.77$\pm$0.15
& 0.58$\pm$0.45
& 0.76$\pm$0.16 \\
\hline
\end{tabular}
\end{table}
%
\section{Conclusion}
We presented \mammolife, a patient-level multimodal framework for post-radiotherapy outcome prediction using paired longitudinal mammograms and selected clinical variables. Experimental results on an in-house cohort demonstrated that integrating longitudinal mammographic representations with selected clinical variables improves patient-level outcome prediction over image-only modeling. Future work will validate \mammolife{} on larger independent cohorts and investigate its generalizability across diverse clinical settings.

\begin{credits}
\subsubsection{\discintname}
The authors have no competing interests to declare that are relevant to the content of this article.
\end{credits}
%

\bibliographystyle{splncs04}
\bibliography{References}
\end{document}